\documentclass[conference]{IEEEtran}  
\usepackage{times}
\usepackage{epsfig}
\usepackage{graphicx}
\usepackage{amsmath}
\usepackage{amssymb}

\usepackage{multirow}
\usepackage{makecell}
\usepackage{color}
\usepackage{arydshln}

\usepackage[pagebackref=true,breaklinks=true,colorlinks,bookmarks=false]{hyperref}

\begin{document}

\title{Multi-view Self-Constructing Graph Convolutional Networks with Adaptive Class Weighting Loss for Semantic Segmentation}

\author{Qinghui Liu$^{1,2}$, Michael Kampffmeyer$^{2}$, 
Robert Jenssen$^{2,1}$, Arnt-B{\o}rre Salberg$^{1}$ \\
$^{1}$Norwegian Computing Center, Oslo, NO-0314, Norway\\

$^{2}$UiT Machine Learning Group, UiT the Arctic University of Norway, Troms{\o}, Norway\\
{\tt\small liu@nr.no}, {\tt\small \{michael.c.kampffmeyer, robert.jenssen\}@uit.no}, {\tt\small salberg@nr.no}
}

\maketitle

\begin{abstract}
We propose a novel architecture called the Multi-view Self-Constructing Graph Convolutional Networks (MSCG-Net) for semantic segmentation. Building on the recently proposed Self-Constructing Graph (SCG) module, which makes use of learnable latent variables to self-construct the underlying graphs directly from the input features without relying on manually built prior knowledge graphs, we leverage multiple views in order to explicitly exploit the rotational invariance in airborne images. We further develop an adaptive class weighting loss to address the class imbalance. We demonstrate the effectiveness and flexibility of the proposed method on the Agriculture-Vision challenge dataset and our model achieves very competitive results (0.547 mIoU) with much fewer parameters and at a lower computational cost compared to related pure-CNN based work. 
Code will be available at: github.com/samleoqh/MSCG-Net
\end{abstract}


\section{Introduction}
Currently, the end-to-end semantic segmentation models are mostly inspired by the idea of fully convolutional networks (FCNs)~\cite{long2015fully} that generally consist of an encoder-decoder architecture. To achieve higher performance, CNN-based end-to-end methods normally rely on deep and wide multi-scale CNN architectures to create a large receptive field in order to obtain strong local patterns, but also capture long range dependencies between objects of the scene. However, this approach for modeling global context relationships is highly inefficient and typically requires a large number of trainable parameters, considerable computational resources, and large labeled training datasets. 

\begin{figure}[tphb!]
 \centering
  \includegraphics[width=0.50\textwidth]{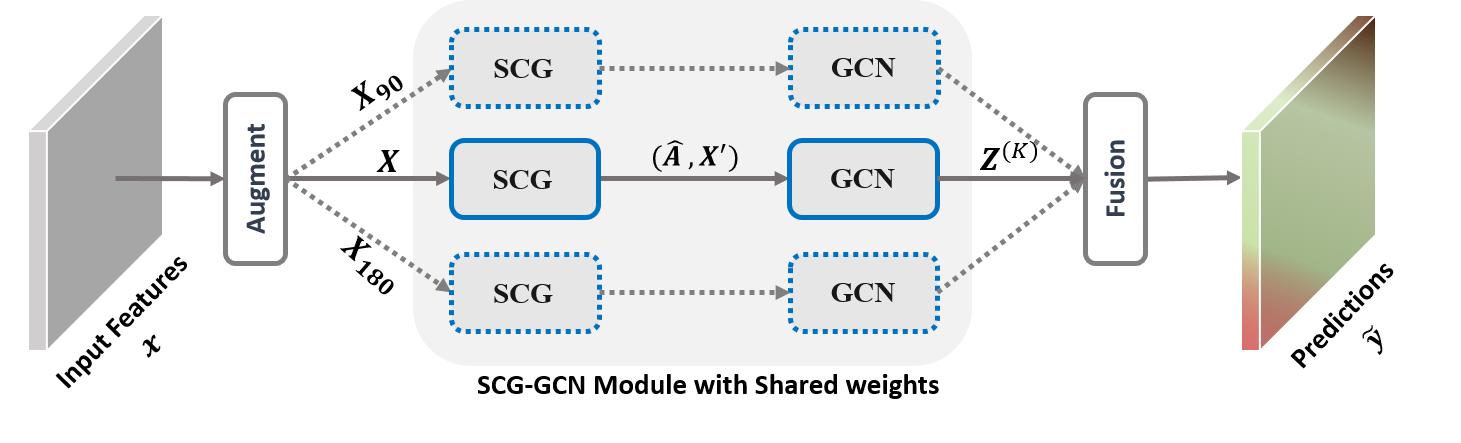} 
      \caption{Overview of the MSCG-Net. The Self-Constructing Graph module (SCG) learns to transform a 2D feature map into a latent graph structure and assign pixels ($X^\prime$) to the vertices of the graph ($\hat{A}$). The Graph Convolutional Networks (GCN) is then exploited to update the node features ($Z^{(K)}$, $K$ here denotes the number of layer of GCN) along the edges of graph. The combined module of SCG and GCN (SCG-GCN) can takes augmented multi-view input features ($X$, $X_{90}$ and $X_{180}$, where the index indicates degree rotation) and finally the updated multi-view representations are fused and projected back onto 2D maps.}
  \label{fig:overall_scg_net}
\end{figure}

Recently, graph neural networks (GNNs) \cite{ bronstein2017geometric} and Graph Convolutional Networks (GCNs) \cite{kipf2016semi} 
have received increasing attention and have been applied to, among others, image classification~\cite{knyazev2019image}, few-shot and zero-shot classification~\cite{kampffmeyer2019rethinking}, point clouds classification~\cite{wang2019dynamic} and semantic segmentation~\cite{liang2018symbolic}. However, these approaches are quite sensitive to how the graph of relations between objects is built and previous approaches commonly rely on manually built graphs based on prior knowledge~\cite{liang2018symbolic}.
In order to address this problem and learn a latent graph structure directly from 2D feature maps for semantic segmentation, the Self-Constructing Graph module (SCG)~\cite{liu2020self} was recently proposed and has obtained promising results.

In this work, we extend the SCG to explicitly exploit the rotation invariance in airborne images, by extending it to consider multiple views. More specifically, we augment the input features to obtain multiple rotated views and fuses the multi-view global contextual information before projecting the features back onto the 2-D spatial domain. We further propose a novel adaptive class weighting loss that addresses the issue of class imbalance commonly found in semantic segmentation datasets. Our experiments demonstrate that the MSCG-Net achieves very robust and competitive results on the Agriculture-Vision challenge dataset, which is a subset of the Agriculture-Vision dataset~\cite{chiu2020agriculturevision}. 

The rest of the paper is organized as follows. In the method Section~\ref{method}, we present the methodology in details. Experimental procedure and evaluation of the proposed method is performed in Section~\ref{exp}. Finally in Section~\ref{concl}, we draw conclusions.

\section{Methods}
\label{method}
In this section, we briefly present graph convolutional networks and the self-constructing graph (SCG) approach that are the foundation of our proposed model, before presenting our end-to-end trainable Multi-view SCG-Net (MSCG) for semantic labeling tasks with the proposed adaptive class weighting loss. 

\subsection{Graph Convolutional Networks}
Graph Convolutional Networks (GCNs)~\cite{kipf2016semi} are neural networks designed to operate on and extract information from graphs and were originally proposed for the task of semi-supervised node classification. $G = (A, X)$ denotes an undirected graph with $n$ nodes, where $A \in \mathbb{R}^{n \times n}$ is the adjacency matrix and $X \in \mathbb{R}^{n \times d}$ is the feature matrix. At each layer, the GCN aggregates information in one-hop neighborhoods, more specifically, the representation at layer $l+1$ is computed as
\begin{equation} \label{eq:gcn2}
Z^{(l+1)}=\sigma\left(\hat{A} Z^{(l)} \theta^{(l)}\right) \; ,
\end{equation} 
where $\theta^{(l)} \in \mathbb{R}^{d \times f}$ are the weights of the GCN, $Z^{(0)}=X$, and
$\hat{A}$ is the symmetric normalization of $A$ including self-loops~\cite{liu2020self}:
\begin{equation} \label{eq:norm}
\hat{A}=D^{-\frac{1}{2}}(A+I)D^{\frac{1}{2}} \; ,
\end{equation}
where $D_{ii} = \sum_{j}(A+I)_{ij}$ is the degree matrix, $I$ is the identity matrix, and $\sigma$ denotes the non-linearity function (e.g. $ReLU$). 

Note, in the remainder of the paper, we use $Z^{(K)} = \operatorname{GCN}(A, X)$ to denote the activations after a $K$-layer GCN. However, in practice the $\operatorname{GCN}$ could be replaced by alternative graph neural network modules that perform $K$ steps of message passing based on some adjacency matrix $A$ and input node features $X$.

\subsection{Self-Constructing Graph}
The Self-Constructing Graph (SCG) module~\cite{liu2020self} allows the construction of undirected graphs, capturing relations across the image, directly from feature maps, instead of relying on prior knowledge graphs. It has achieved promising performance on semantic segmentation tasks in remote sensing and is efficient with respect to the number of trainable parameters, outperforming much larger models.
It is inspired by variational graph auto-encoders~\cite{kipf2016variational}. 
A feature map $X \in \mathbb{R}^{h \times w \times d}$ consisting of high-level features, commonly produced by a CNN, is converted to a graph $G = (\hat{A}, X^{\prime})$. $X^{\prime} \in \mathbb{R}^{n \times d}$ are the node features, where $n = h^{\prime} \times w^{\prime}$ denotes the number of nodes and where $(h^\prime \times w^\prime) \leq (h \times w)$. Parameter-free pooling operations, in our case adaptive average pooling, are employed to reduce the spatial dimensions of $X$ to $h^\prime$ and $w^\prime$, followed by a reshape operation to obtain $X^\prime$. $\hat{A} \in \mathbb{R}^{n \times n}$ is the learned weighted adjacency matrix.

The SCG module learns a mean matrix $\boldsymbol{\mu} \in \mathbb{R}^{n \times c}$
and a standard deviation matrix $\boldsymbol{\sigma} \in \mathbb{R}^{n \times c}$ of a Gaussian using two single-layer convolutional networks. 
Note, following convention with variational autoencoders~\cite{kingma2013auto}, the output of the model for the standard deviation is $\log(\sigma)$ to ensure stable training and positive values for $\sigma$. With help of reparameterization, the latent embedding $Z$ is $\textit{Z} \leftarrow \boldsymbol{\mu} + \boldsymbol{\sigma} \cdot  \boldsymbol{\varepsilon}$ \;
where $\varepsilon \in \mathbb{R}^{N^{\prime} \times C}$ is an auxiliary noise variable and initialized from a standard normal distribution ($\boldsymbol{\varepsilon} \sim \operatorname{N}(0,I)$). A centered isotropic multivariate Gaussian prior distribution is used to regularize the latent variables, by minimizing a Kullback-Leibler divergence loss 
\begin{equation}\label{eq:klloss}
\mathcal{L}_{kl} = -\frac{1}{2n} \sum_{i=1}^{n}\left(1+\log \left(\sigma_{i} \right)^{2}-\mu_{i}^{2}-\sigma_{i}^{2}\right) \; .
\end{equation}

Based on the learned embeddings, $A^\prime$ is computed as $A^{\prime} = \operatorname{ReLU}(Z Z^{T})$, where $A_{ij}^{\prime} > 0$ indicates the presence of an edge between node $i$ and $j$. 

Liu et al.~\cite{liu2020self} further introduce a diagonal regularization term 
\begin{equation}\label{eq:dlloss}
  \mathcal{L}_{dl} = - \frac{\gamma}{n^2} \sum_{i=1}^{n} \operatorname{log} (\left|A_{ii}^\prime\right|_{[0,1]} + \epsilon)  \; ,
\end{equation}
where $\gamma$ is defined as $$\gamma = \sqrt{1 + \frac{n}{ \sum_{i=1}^{n} (A_{ii}^\prime) + \epsilon}}$$ and a diagonal enhancement approach
\begin{equation}\label{eq:enhance}
    A^\star = A^\prime + \gamma \cdot \operatorname{diag}(A^\prime)
\end{equation}
to stabilize training and preserve local information.

The symmetric normalized $\hat{A}$ that SCG produces and that will be the input to later graph operations is computed as
\begin{equation}\label{eq:normenhance}
\hat{A} = D^{-\frac{1}{2}}\left(A^\star+I\right)D^{\frac{1}{2}} \; .
\end{equation}

The SCG further produces an adaptive residual prediction $\boldsymbol{\hat{y}} = \gamma \cdot \boldsymbol{\mu} \cdot (1 - {\log \boldsymbol{\sigma}}) $, which is used to refine the final prediction of the network after information has been propagated along the graph. 

\subsection{The MSCG-Net} \label{archit}

\begin{figure*}[thpb!]
 \centering
  \includegraphics[width=0.95\textwidth]{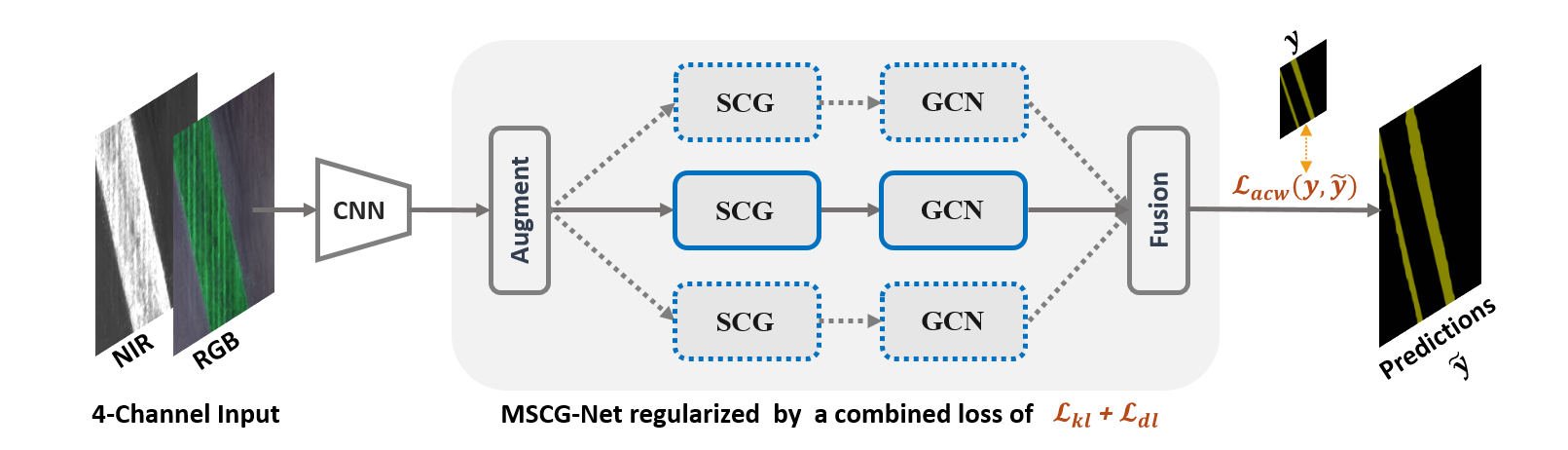} 
  \caption{Model architecture of MSCG-Net for semantic labeling includes the CNN-based feature extractor (e.g. customized Se\_ResNext50\_32x4d taking 4-channel input and output 1024-channel in this work), SCG module taking 3-view (augment the original input by $90^{\circ} \text{and} 180^{\circ}$) inputs and K-layer GCNs (K=2 in this work), the Fusion block merging 3-view outputs together, the fused output is projected and upsampled back to 2D maps for final prediction.}
  \label{fig:mscg_net}
\end{figure*}

We propose a so-called Multi-view SCG-Net (MSCG-Net) to extend the vanilla SCG and GCN modules by considering multiple rotated views in order to obtain and fuse more robust global contextual information in airborne images in this work. Fig.~\ref{fig:mscg_net} shows an illustration of the end-to-end MSCG-Net model for semantic labeling tasks. The model architecture details are shown in Table~\ref{tab:net-config}. 
We first augment the features ($X$) learned by a backbone CNN network to multiple views ($X_{90}  \text{ and } X_{180}$) by rotating the features. The employed SCG-GCN module then outputs multiple predictions: $(\hat{\boldsymbol{y}}, Z^{(2)}),  (\hat{\boldsymbol{y}}_{90}, Z_{90}^{(2)}),  (\hat{\boldsymbol{y}}_{180}, Z_{180}^{(2)})$ with different rotation degrees (the index indicates the degree of rotation). The fusion layer merges all the predictions together by reversed rotations and element-wise additions as shown in Table~\ref{tab:net-config}. Finally, the fused outputs are projected and up-sampled back to the original 2-D spatial domain. 

\begin{table*}[htbp]
\begin{center}
    \begin{tabular}{c|c|c} \hline 
    \textbf{Layers} & \textbf{Outputs} & \textbf{Sizes} \\
    \hline 
    CNN & $X$ & $\frac{h}{16} \times \frac{w}{16} \times 1024$  \\ \hdashline
    Augment & $(X,X_{90},X_{180})$ &  $3 \times (\frac{h}{16} \times \frac{w}{16} \times 1024$) \\ \hdashline
    SCG & $(\hat{A},X^\prime, \hat{\boldsymbol{y}}), (\hat{A}_{90},X_{90}^\prime, \hat{\boldsymbol{y}}_{90}), (\hat{A}_{180},X_{180}^\prime, \hat{\boldsymbol{y}}_{180})$ &  $3 \times [(n\times n), (n\times 1024), (n\times c)]$  \\ \hdashline
    $\operatorname{GCN}^1$ & $(Z^{(1)}, Z_{90}^{(1)}, Z_{180}^{(1)})$ &   $3 \times (n\times d)$      \\ \hdashline
    $\operatorname{GCN}^2$ & $(Z^{(2)}, Z_{90}^{(2)}, Z_{180}^{(2)})$ &   $3\times (n\times c)$      \\ \hdashline
    Fusion & $(\hat{\boldsymbol{y}} + Z^{(2)})\oplus(\hat{\boldsymbol{y}}_{90} + Z_{90}^{(2)})_{r90}\oplus(\hat{\boldsymbol{y}}_{180} + Z_{180}^{(2)})_{r180}$ &   $(n \times c) \longrightarrow (\frac{h}{16} \times \frac{w}{16} \times c)$  \\ \hdashline
    Projection & $\Tilde{\boldsymbol{y}}$ &   $h\times w \times c $  \\
    \hline
    \end{tabular}%
  \label{tab:net-config}%
  \end{center}
  \caption{MSCG-Net Model Details with one sample of input image size of $h\times w \times 4$. Note: $\oplus$ denotes an element-wise addition, the index (i.e. $90$, $180$) indicates the rotated degree, while $r90$ and $r180$ denote the reversed rotation degrees.}
\end{table*}%

We utilize the first three bottleneck layers of a pretrained Se\_ResNext50\_32x4d or Se\_ResNext101\_32x4d~\cite{hu2018squeeze} as the backbone CNN to learn the high-level representations. The output size of the CNN is $\frac{h}{16} \times \frac{w}{16} \times 1024$. Note that, we duplicate the weights corresponding to the Red channel of the pretrained input convolution layer in order to take NIR-RGB 4 channels in the backbone CNN, and GCNs (Equation~\ref{eq:gcn2}) are used in our model. We utilize ReLU activation and batch normalization only for the first layer GCN. Note, we set $n= 32^2$ and $d=128$ in this work, and $c$ here is equal to the number of classes, such that $c = 7$ for the experiments performed in this paper.

\subsection{Adaptive Class Weighting Loss} \label{adloss}
    The distribution of the classes is highly imbalanced in the dataset (e.g. most pixels in the images belongs to the background class and only few belong to classes such as planter skip and standing water). To address this problem, most existing methods make use of weighted loss functions with pre-computed class weights based on the pixel frequency of the entire training data~\cite{kampffmeyer2016semantic} to scale the loss for each class-pixel according to the fixed weight before computing gradients. In this work, we introduce a novel class weighting method based on iterative batch-wise class rectification, instead of pre-computing the fixed weights over the whole dataset. 
    
    The proposed adaptive class weighting method is derived from median frequency balancing weights~\cite{kampffmeyer2016semantic}. We first compute the pixel-frequency of class $j$  over all the past training steps as follows 
\begin{equation}\label{freq}
    f^{t}_j = \frac{\hat{f}^{t}_j + (t-1)*f^{t-1}_j}{t}\;.
\end{equation}
where, $t \in \{1,2,..., \infty\}$ is the current training iteration number, $\hat{f}^{t}_j$ denotes the pixel-frequency of class $j$ at the current $t\text{-th}$ training step that can be computed as $\frac{\operatorname{SUM}(y_j)}{\sum_{j \in C} \operatorname{SUM}(y_j)}$, and $f^{0}_j = 0$.

The iterative median frequency class weights can thus be computed as
\begin{equation}\label{eq:mfb}
\textit{w}^{t}_j =\frac{\text{median} (\{f^{t}_j | j\in C \})}{f^{t}_j + \epsilon}\;.
\end{equation}
here, $C$ denotes the number of labels ($7$ in this paper), and $\epsilon = 10^{-5}$. 

Then we normalize the iterative weights with adaptive broadcasting to pixel-wise level such that
\begin{equation}\label{eq:acw}
\Tilde{w}_{ij} = \frac{\textit{w}^{t}_j }{\sum_{j \in C}(\textit{w}^{t}_j)} * (1+y_{ij}+\Tilde{y}_{ij})\;,
\end{equation}
where $\Tilde{y}_{ij} \in (0, 1)$ and $y_{ij} \in \{0, 1\}$ denote the $ij$-th prediction and the ground-truth of class $j$ separately in the current training samples.

In addition, instead of using traditional cross-entropy function which focuses on positive samples, we introduce a positive and negative class balanced function (PNC) which is defined as
\begin{equation}\label{eq:pn_loss}
\boldsymbol{p} = \boldsymbol{e} - \operatorname{log}\left(\frac{1 - \boldsymbol{e}}{1 + \boldsymbol{e}} \right)\;,
\end{equation}
where $\boldsymbol{e} = \left(\boldsymbol{y} -  \Tilde{\boldsymbol{y}}\right)^{2}$.

Building on the dice coefficient \cite{milletari2016v} with our adaptive class weighting PNC function, we develop an adaptive multi-class weighting (ACW) loss function for multi-class segmentation tasks
 \begin{equation}\label{eq:acw_loss}
 \mathcal{L}_{acw} = \frac{1}{\left|{Y}\right|}  \sum_{i \in Y} \sum_{j \in C} \Tilde{w}_{ij}*p_{ij} - \operatorname{log}\left( \text{MEAN} \{d_j | j\in C\}\right)  \;,
 \end{equation}
where $Y$ contains all the labeled pixels and $d_j$ is the dice coefficient given as
\begin{equation}\label{eq:dice}
    d_j = \frac{2 \sum_{i \in Y} y_{ij} \Tilde{y}_{ij}}{\sum_{i \in Y} y_{ij} + \sum_{i \in Y} \Tilde{y}_{ij}}\; .
\end{equation}

The overall cost function of our model, with a combination of two regularization terms $\mathcal{L}_{kl}$ and $\mathcal{L}_{dl}$ as defined in the equations \ref{eq:klloss} and \ref{eq:dlloss},  is therefore defined as
\begin{equation}\label{eq:cost_function}
    \mathcal{L} \gets \mathcal{L}_{acw} + \mathcal{L}_{kl} + \mathcal{L}_{dl} \; .
\end{equation}

\section{Experiments and results}
\label{exp}
We first present the training details and report the results. We then conduct an ablation study to verify the effectiveness of our proposed methods. 
\subsection{Dataset and Evaluation}
We train and evaluate our proposed method on the Agriculture-Vision challenge dataset, which is a subset of the Agriculture-vision dataset~\cite{chiu2020agriculturevision}. The challenge dataset consists of $21,061$ aerial farmland images captured throughout 2019 across the US. Each image contains four 512x512 color channels, which are RGB and Near Infra-red (NIR). Each image has a boundary map that indicates the region of the farmland, and a mask that indicates valid pixels in the image. Seven types of annotations are included: Background, Cloud shadow, Double plant, Planter skip, Standing Water, Waterway and Weed cluster. Models are evaluated on the validation set with $4,431$ NIR-RGB images segmentation pairs, while the final scores are reported on the test set with $3,729$ images. The mean Intersection-over-Union (mIoU) is used as the main quantitative evaluation metric. Due to the fact that some annotations may overlap in the dataset, for pixels with multiple labels, a prediction of either label will be counted as a correct pixel classification for that label. 

\subsection{Training details}
We use backbone models pretrained on ImageNet in this work. We randomly sample patches of size $512\times 512$ as input and train it using mini batches of size $10$ for the MSCG-Net-50 model and size $7$ for the MSCG-Net-101 model. The training data (containing 12901 images) is sampled uniformly and randomly flipped (with probability 0.5) for data augmentation and shuffled for each epoch. 

According to our best practices, we first train the model using Adam~\cite{KingmaB14adam} combined with Lookahead~\cite{zhang2019lookahead} as the optimizer for the first 10k iterations and then change the optimizer to SGD in the remaining iterations with weight decay $2 \times 10^{-5}$ applied to all learnable parameters except biases and batch-norm parameters. We also set $2 \times LR$ to all bias parameters compared to weight parameters. Based on our training observations and empirical evaluations, we use initial LRs of $\frac{1.5 \times 10^{-4}}{\sqrt{3}}$ and $\frac{2.18 \times 10^{-4}}{\sqrt{3}}$ for MSCG-Net-50 and MSCG-Net-101 separately, and also apply cosine annealing scheduler that reduces the LR over epochs. All models are trained on a single NVIDIA GeForce GTX 1080Ti. It took roughly 10 hours to train our model for 25 epochs with batch size 10 over $12,901$ NIR-RGB training images. 

\subsection{Results}

\begin{table*}[hptb!]
\begin{center}
\setlength{\tabcolsep}{2pt}
\resizebox{\textwidth}{!}{
\begin{tabular}{|c|p{12mm}|p{18mm}p{22mm}p{20mm}p{20mm}p{23mm}p{16mm}p{20mm}|p{12mm}|} \hline
    \textbf{Models} & $\textbf{mIoU}$ & \textbf{Background} & \textbf{Cloud shadow} & \textbf{Double plant} & \textbf{Planter skip} & \textbf{Standing water}  & \textbf{Waterway}  & \textbf{Weed cluster} & $\textbf{mIoU}^{\ast}$ \\  \hline \hline
     $\textbf{MSCG-Net-50}$ & 0.547  & 0.780  & \textbf{0.507}  & 0.466 & \textbf{0.343}  & \textbf{0.688}  & 0.513 & \textbf{0.530}& 0.508\\  
     $\textbf{MSCG-Net-101}$ & \textbf{0.550}  & 0.798  & 0.448  & \textbf{0.550} & 0.305  & 0.654  & \textbf{0.592} & 0.506 & \textbf{0.509} \\ \hline  
     $\textit{Ensemble-TTA}$ & 0.599  & 0.801  & 0.503 & 0.576  & 0.520  & 0.696 & 0.560 & 0.538 & 0.566\\ \hline
    \end{tabular}
} 
\end{center}

\caption{mIoUs and class IoUs of our models on Agriculture-Vision test set. Note: mIoU is the mean IoU over all 7 classes while $\text{mIoU}^{\ast}$ is over 6-class without the background, and Ensemble-TTA denotes the two models ensemble (MSCG-Net-50 with MSCG-Net-101) combined with TTA methods \cite{Liu_2020_Journal}.}
\label{tab:leaderboard_scores}%
\end{table*}

\begin{figure}[htpb!]
 \centering
  \includegraphics[width=0.5\textwidth]{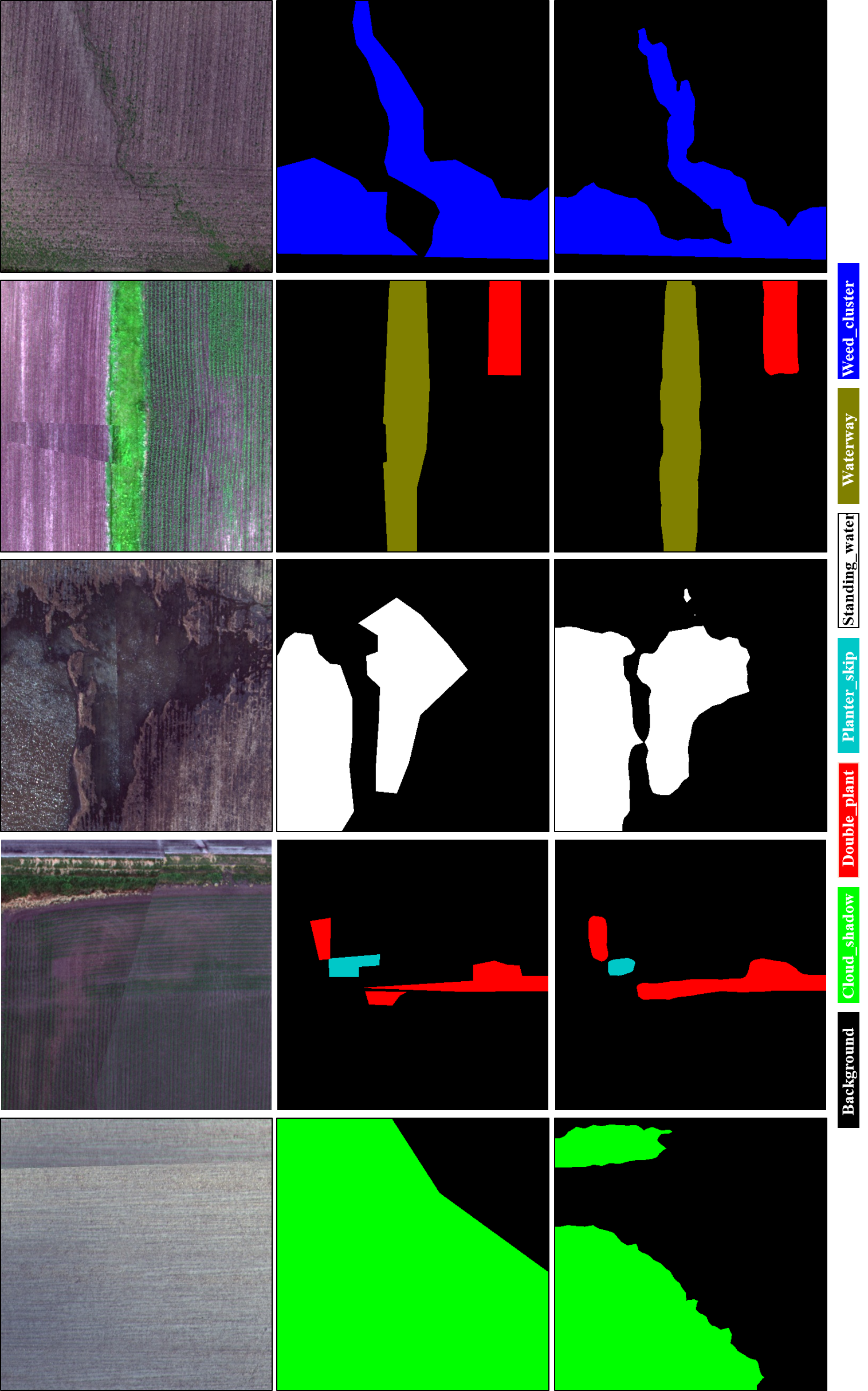}
  \caption{Segmentation results on validation data. From the left to right, the input images, the ground truths and the predictions of our trained models.}
  \label{fig:test}
\end{figure}

We evaluated and tested our trained models on the validation sets and the hold out test sets with just single feed-forward inference without any test time augmentation (TTA) or models ensemble. However, we do include results for a simple two-model ensemble (MSCG-Net-50 together with MSCG-Net-101) with TTA for completeness. The test results are shown in Table~\ref{tab:leaderboard_scores}. Our MSCG-Net-50 model obtained very competitive performance with 0.547 mIoU with very small training parameters ($9.59$ million) and has low computational cost ($18.21$ Giga FLOPs with input size of $4 \times 512 \times 512$), resulting in fast training and inference performance on both CPU and GPU as shown in Table~\ref{tab:parameters}. A qualitative comparisons of the segmentation results from our trained models and the ground truths on the validation data are shown in Fig.~\ref{fig:test}.

\begin{table*}[hptb!]
\begin{center}
\begin{tabular}{c|c|p{16mm}p{12mm}p{25mm}} \hline 
\textbf{Models} & \textbf{Backbones} & \textbf{Parameters} \newline (Million)& \textbf{FLOPs} \newline (Giga) & \textbf{Inference time} \newline (ms - CPU/GPU)   \\ \hline  \hline

$\text{MSCG-Net-50}$ & $\text{Se\_ResNext50}$ & 9.59  &18.21 & {522} / 26 \\
$\text{MSCG-Net-101}$& $\text{Se\_ResNext101}$ & 30.99  &37.86 & 752 / 45  \\ 
\hline
\end{tabular}
\label{tab:parameters}%
\end{center}
 \caption{Quantitative Comparison of parameters size, FLOPs (measured on input image size of $4 \times 512 \times 512$), Inference time on CPU and GPU separately.}
\end{table*}

\begin{table}
\begin{center}
\begin{tabular}{c|c|c|c|c}
\hline
& \multicolumn{3}{c|}{Configurations} \\ \cline{2-4} 
\multirow{-2}{*}{Models} 
    &  \thead{Multi-view} 
        & \thead{Dice loss} 
            & \thead{ACW loss} & \multirow{-2}{*}{mIoU} \\ \hline
SCG-dice  &            & \checkmark &            &  0.456       \\ \hline
SCG-acw  &            &            & \checkmark &  0.472          \\ \hline
MSCG-dice  & \checkmark & \checkmark &            & 0.516 \\ \hline
MSCG-acw  & \checkmark &           &  \checkmark & 0.527           \\ \hline
\end{tabular}
\end{center}
\caption{Ablation study of our proposed network. Note that, for simplicity, we fixed the learning high-parameters and the backbone encoder, and mIoU is evaluated on validation set without considering overlapped annotations.}
\label{tab:ablation}
\end{table}

\begin{figure*}[htpb!]
 \centering
  \includegraphics[width=1\textwidth]{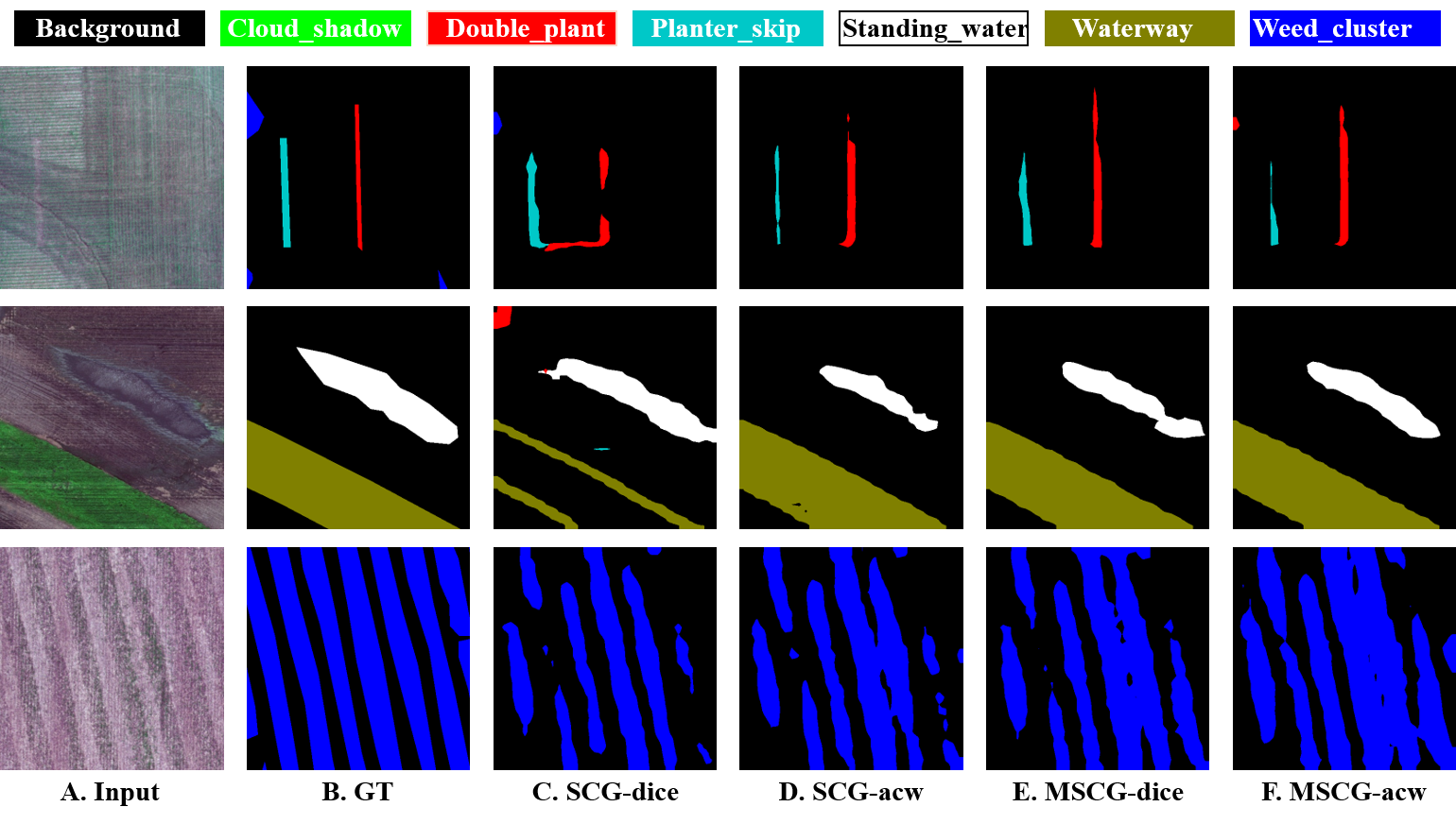}
  \caption{Segmentation results using different models. From the left to right, the input images, the ground truths and SCG-Net with dice loss, SCG-Net with ACW loss, MSCG-Net with dice loss, and MSCG-Net with ACW loss.}
  \label{fig:ablation}
\end{figure*}

\subsection{Ablation studies}
\textbf{Effect of the Multi-view.}
To investigate how the multiple views help, we report the results of the single-view models and the multi-view models trained with both Dice loss and ACW loss in Table \ref{tab:ablation}. Note that, for simplicity, we fixed the backbone encoder as Se\_ResNext50 and other training parameters (e.g. learning rate, decay police, and so on.). Also, the mIoUs are computed on the validation set without considering multiple labels. The results suggest that multiple views could improve the overall performance from $0.456$ to $0.516$ ($+6\%$) mIoU when using Dice loss, and from $0.472$ to $0.527$ ($+5.5\%$) with the proposed ACW loss. 

\textbf{Effect of the ACW loss.} As shown in Table~\ref{tab:ablation}, we note that for the single-view models, the overall performance can be improved from $0.456$ to $0.472$ ($+1.6\%$) mIoU. For the multi-view models, the performance improved $+1.1\%$, increasing from $0.516$ to $0.527$. 
Compared to the single-view model SCG-Net with Dice loss, which was proposed in~\cite{liu2020self} and achieved state-of-the-art performance on a commonly used segmentation benchmark dataset, our Multi-view MSCG-Net model with ACW loss achieved roughly $+7.1\%$ higher mIoU accuracy. We show some qualitative results in Fig.~\ref{fig:ablation} that illustrate the proposed multi-view model and the adaptive class weighting method and show that they help to produce more accurate segmentation results for both larger and smaller classes.

\section{Conclusions} \label{concl}
In this paper, we presented a multi-view self-constructing graph convolutional network (MSCG-Net) to extend the SCG module which makes use of learnable latent variables to self-construct the underlying graphs, and to explicitly capture multi-view global context representations with rotation invariance in airborne images.  We further developed a novel adaptive class weighting loss that alleviates the issue of class imbalance commonly found in semantic segmentation datasets. On the Agriculture-Vision challenge dataset, our MSCG-Net model achieves very robust and competitive results, while making use of fewer parameters and being computationally more efficient. 

\section*{Acknowledgments}
This work is supported by the foundation of the Research Council of Norway under Grant 220832.

{\small
\bibliographystyle{ieee_fullname}
\bibliography{my}
}

\end{document}